\documentclass[sigconf, screen]{acmart}
\usepackage{soul}
\usepackage{mwe}
\usepackage{caption}

\AtBeginDocument{%
  }

\setcopyright{iw3c2w3}
\copyrightyear{2025}
\acmYear{2025}
\begin{document}
\acmConference[CONSEQUENCES Workshop @ RecSys '25]{The 19th ACM Conference on Recommender Systems}{September 22, 2025}{Prague, Czech Republic}

\title{Direct Profit Estimation Using Uplift Modeling under Clustered Network Interference}

\author{Bram van den Akker}
\email{bram.vandenakker@booking.com}
\affiliation{%
  \institution{Booking.com}
  \city{Amsterdam}
  \country{The Netherlands}
}
\begin{abstract}
    Uplift modeling is a key technique for promotion optimization in recommender systems, but standard methods typically fail to account for interference, where treating one item affects the outcomes of others. This violation of the Stable Unit Treatment Value Assumption (SUTVA) leads to suboptimal policies in real-world marketplaces. Recent developments in interference-aware estimators such as Additive Inverse Propensity Weighting (AddIPW) have not found their way into the uplift modeling literature yet, and optimising policies using these estimators is not well-established. This paper proposes a practical methodology to bridge this gap. We use the AddIPW estimator as a differentiable learning objective suitable for gradient-based optimization. We demonstrate how this framework can be integrated with proven response transformation techniques to directly optimize for economic outcomes like incremental profit. Through simulations, we show that our approach significantly outperforms interference-naive methods, especially as interference effects grow. Furthermore, we find that adapting profit-centric uplift strategies within our framework can yield superior performance in identifying the highest-impact interventions, offering a practical path toward more profitable incentive personalization.
\end{abstract}


\received{7 August 2025}
\received[revised]{1 September 2025}

\maketitle

\begin{figure*}[h!]
    \centering
    \begin{minipage}{0.48\textwidth}
        \centering
        \includegraphics[width=\textwidth]{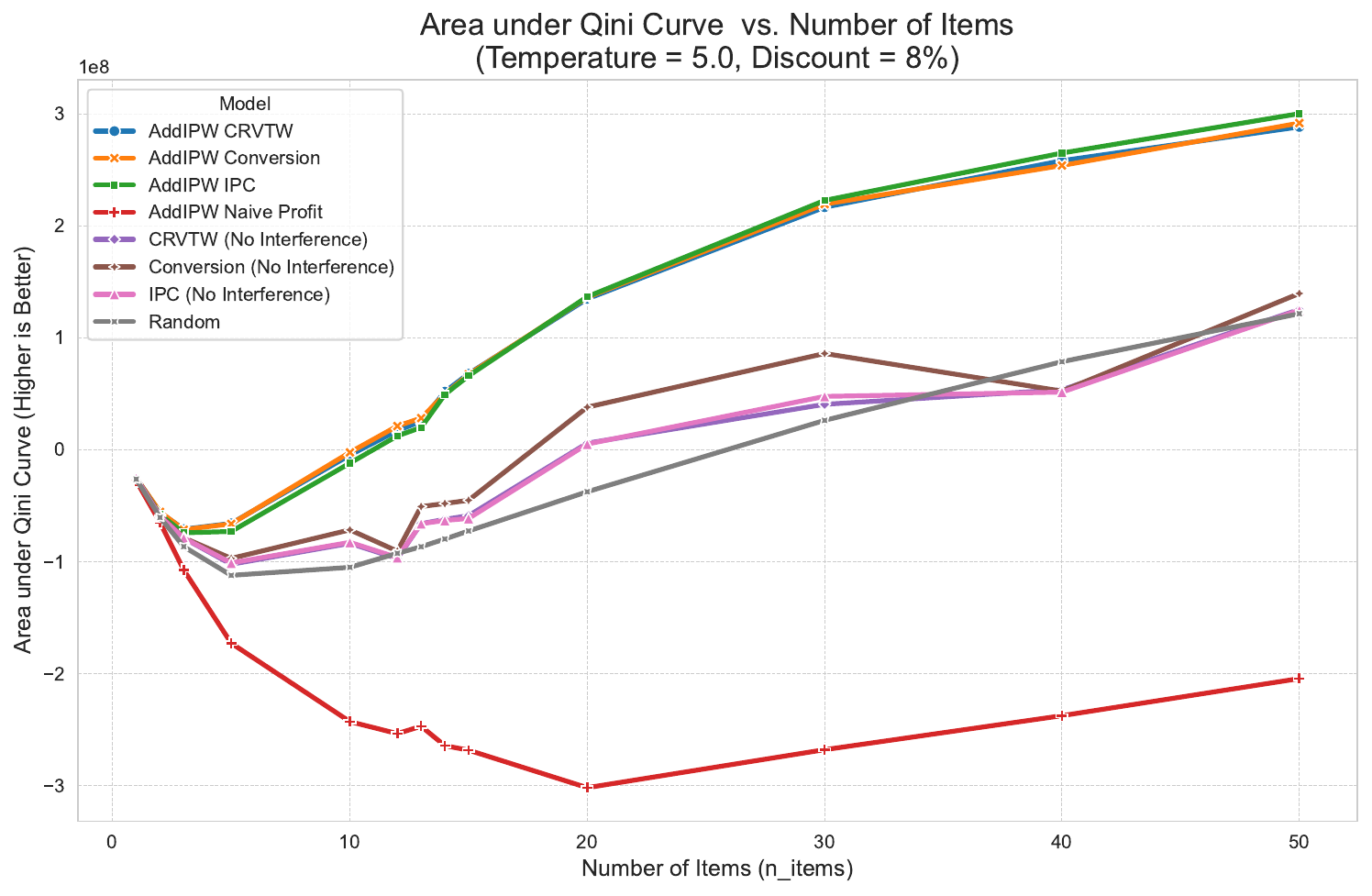}
        \captionof{figure}{\textit{The Area under the Qini curve}, comparing various estimation strategies. As the cluster size (n\_items) grows, the advantage of using AddIPW as a weighting strategy increases relative to naive approaches.}\label{fig:aucvsclusters}
    \end{minipage}
    \hfill
    \begin{minipage}{0.48\textwidth}
        \centering
        \includegraphics[width=\textwidth]{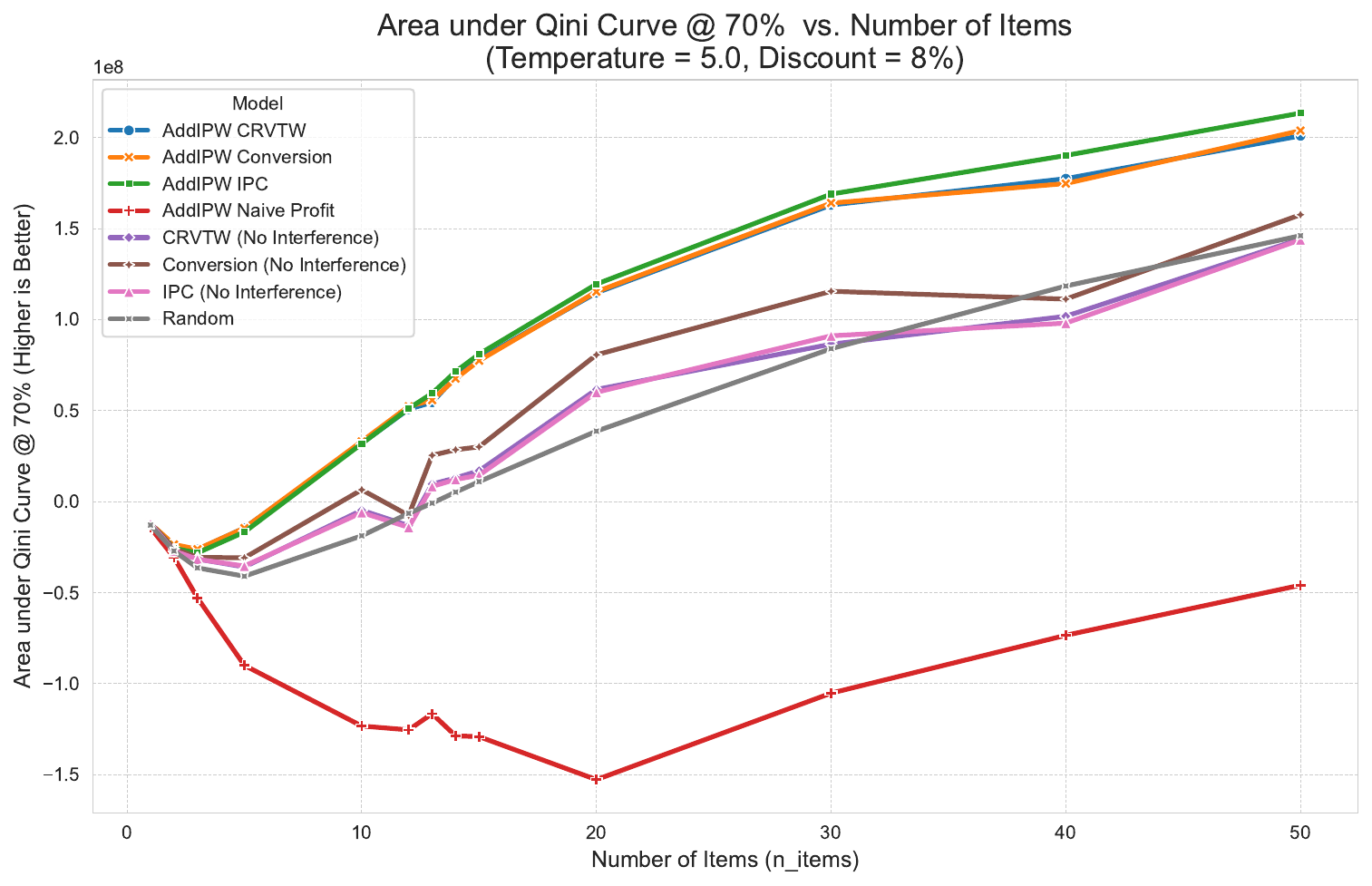}
        \captionof{figure}{\textit{The Area under the Qini curve of the first 70\% of items treated}, comparing various estimation strategies.}\label{fig:auc70vscluster}
    \end{minipage}
\end{figure*}

\section{Introduction}
Recommender systems are increasingly understood as decision-making engines that determine not only what content to display, but also how and in what order. Beyond item ranking~\cite{agarwal2019general, oosterhuis2021unifying}, a critical function of these systems is deciding whether to accompany a recommendation with a promotion, such as a discount or coupon. This task of personalizing interventions can be solved using uplift modeling\cite{radcliffe2007using}, which aims to prioritize the treatment of interventions by estimating the causal effect of a treatment on individual behavior to optimize for economic outcomes like profit or return on investment~\cite{gubela2020response, proencca2023incremental, zhou2023direct}. This literature commonly relies on the Stable Unit Treatment Value Assumption~\citep{rubin1980randomization}, which implies that treating one unit has no influence on the outcome of other units. 

However, in practice, such interventions do not occur in a vacuum. In real-world marketplaces, they create complex interference effects where treating one item influences the outcomes of others, for example by cannibalizing sales. A common approach to managing this complexity is the clustered network interference assumption~\cite{hudgens2008toward}, where effects are contained within a defined group. In an e-commerce setting, for instance, all items shown to a user can be treated as a cluster; promoting one item may affect which of those items the user purchases, but it will not affect the purchasing decisions of other users.

While this assumption provides a tractable framework, learning interference-aware policies remains a challenge. Recent foundational work has provided the necessary tools for evaluation. \cite{zhang2023individualized} introduced the Additive Inverse Propensity Weighting (AddIPW) estimator, an efficient method for learning and evaluating policy outcomes under interference. Subsequently, Karlsson et al. (2025)~\cite{karlsson2025qini} demonstrated how this estimator could be used for the robust evaluation of uplift models via Qini curves. A Qini curve visualizes the incremental gain of a treatment policy versus the cost of its implementation, allowing practitioners to assess a policy's cost-effectiveness across different allocation budgets~\cite{karlsson2025qini}.

This paper bridges the gap from evaluation to learning. While the AddIPW framework~\cite{zhang2023individualized} can be used for policy optimization, its direct application to economic metrics is not straightforward as the estimator lacks an explicit cost component. We propose a methodology to adapt the AddIPW estimator to target real-world economic goals, such as incremental profit. Our methodology integrates AddIPW with existing uplift modeling techniques, like response transformation, to create effective and practical incentive policies in the presence of network interference.

\section{Methodology}
We adopt the notation from recent work on clustered network interference~\cite{karlsson2025qini, zhang2023individualized}. Let there be $n$ independent clusters (e.g., users), indexed by $i=1, ..., n$. Each cluster $i$ contains $M_i$ units (e.g., items), indexed by $j=1, ..., M_i$. For each unit $(i,j)$, we observe pre-treatment covariates $X_{ij}$, a binary treatment $A_{ij} \in \{0,1\}$, and an outcome $Y_{ij}$. The collection of all covariates for a cluster is denoted by $X_i$. A key distinction is that the individual-level outcome is $Y_{ij}$ (e.g., item $j$ was purchased), while the cluster-level outcome, $\overline{Y}_i$, is the average of individual outcomes within that cluster: $\overline{Y}_i = \frac{1}{M_i}\sum_{j=1}^{M_i} Y_{ij}$. The propensity score, $e_j(a|X_i) = P(A_{ij}=a|X_i)$, represents the probability of a unit receiving treatment $a$, conditional on the covariates of the *entire cluster*, thus accounting for potential dependencies in treatment assignment.

We intend to learn an optimal, individualized policy, $\pi(X_{ij})$, for applying incentives in a marketplace setting. For each item $j$ presented to a user $i$ within a cluster, the policy must decide whether to apply a promotion. The goal is to learn this policy from observational or experimental data such that it maximizes a desired economic outcome, like incremental profit, when deployed. A key constraint is that the learning method must be robust to the clustered network interference inherent in such marketplace environments.

\subsection{The AddIPW Estimator}
Before detailing our learning objective, we first introduce the Additive Inverse Propensity Weighting (addIPW) estimator for evaluating a policy $\pi$ under clustered network interference, as proposed by Zhang and Imai ~\cite{zhang2023individualized}. Under an additive outcome model assumption, the value of a policy $V(\pi)$ can be estimated efficiently. The estimator is defined as:

\begin{equation}
    \hat{V}^{\text{addIPW}}(\pi) = \frac{1}{n}\sum_{i=1}^{n}\overline{Y}_{i}\left[\sum_{j=1}^{M_{i}}\left(\frac{\mathbb{I}\{A_{ij}=\pi(X_{ij})\}}{e_{j}(\pi(X_{ij})|X_{i})}-1\right)+1\right]
    \label{eq:addipw_full}
\end{equation}

Unlike traditional IPW estimators that reweight entire clusters and have weights that scale exponentially with cluster size, the addIPW estimator's weights scale linearly, making it substantially more efficient\cite{karlsson2025qini}. 

Our methodology adapts the AddIPW estimator into a learning objective suitable for uplift modeling. The general objective for policy learning with AddIPW is to find the policy $\pi$ that maximizes the estimated policy value:

\begin{equation}
    \hat{\pi} := \arg\max_{\pi \in \Pi} \hat{V}^{addIPW}(\pi)
\end{equation}

To create a learning objective, the part of the estimator that depends on the policy $\pi$ can be isolated. By separating the policy-dependent terms from the constants, the optimization problem simplifies to:

\begin{equation}
    \max_{\pi \in \Pi} \frac{1}{n}\sum_{i=1}^{n}\overline{Y}_{i}\sum_{j=1}^{M_{i}}\left(\frac{\mathbb{I}\{A_{ij}=\pi(X_{ij})\}}{e_{j}(\pi(X_{ij})|X_{i})}\right)
\end{equation}

Expanding the indicator function $\mathbb{I}\{\cdot\}$ for a binary policy $\pi(X_{ij}) \in \{0,1\}$, this is equivalent to:

\begin{equation}
    \max_{\pi \in \Pi} \frac{1}{n}\sum_{i=1}^{n}\overline{Y}_{i}\sum_{j=1}^{M_{i}}\left[ \pi(X_{ij})\frac{A_{ij}}{e_{j}(1|X_{i})} + (1-\pi(X_{ij}))\frac{1-A_{ij}}{e_{j}(0|X_{i})} \right]
\end{equation}

After rearranging, the objective is to maximize:
\begin{equation}
    \max_{\pi \in \Pi} \frac{1}{n}\sum_{i=1}^{n}\overline{Y}_{i}\sum_{j=1}^{M_{i}}\left(\frac{A_{ij}}{e_{j}(1|X_{i})} - \frac{1-A_{ij}}{e_{j}(0|X_{i})}\right)\pi(X_{ij})
    \label{eq:general_objective}
\end{equation}

Here, $\overline{Y}_i$ is the average outcome for cluster $i$. To enable gradient-based learning, the discrete policy decision $\pi(X_{ij})$ is replaced with a continuous, differentiable function $f_\theta(X_{ij}) \in [0,1]$, representing the probability of treatment. This allows the use of models like logistic regression or gradient-boosted decision trees.

To clarify the connection to existing uplift techniques, consider the common case of a Randomized Controlled Trial (RCT) where the treatment assignment probability is uniform, i.e., $e_j(1|X_i) = 0.5$ for all units. This simplifies the per-item weight in Equation \ref{eq:general_objective} to:
$$ \left(\frac{A_{ij}}{0.5} - \frac{1-A_{ij}}{0.5}\right) = 2A_{ij} - 2(1-A_{ij}) = 4A_{ij} - 2 $$
In this specific case, the learning objective becomes maximizing $\sum_{i,j} \overline{Y}_i \cdot (4A_{ij} - 2) \cdot f_\theta(X_{ij})$. The term $\overline{Y}_i \cdot (2A_{ij} - 1)$ can be interpreted as an instance weight for a standard supervised learning problem, but it is important to note this simplified form is only valid for $e_j=0.5$. Our implementation uses the general form from Equation \ref{eq:general_objective}.

\subsection{Adapting AddIPW for Economic Outcomes}
Our primary goal is to optimize for economic outcomes, such as incremental profit, rather than just conversions. A naive approach might be to simply replace the cluster outcome $\overline{Y}_i$ (average conversions) in the AddIPW objective with the average observed cluster profit $\overline{Y}_{p,i} = \overline{Y}_{r,i} - \overline{Y}_{c,i}$, where $\overline{Y}_{r,i}$ is revenue and $\overline{Y}_{c,i}$ is cost. However, this strategy is likely to fail. The model would learn to penalize treated items simply because their observed profit is lower due to the incentive cost, without correctly isolating the \textit{incremental gain} generated by the promotion. This motivates adapting more sophisticated uplift methods that are designed to estimate causal impact.

An effective technique in uplift modeling is the use of a response transformation~\cite{proencca2023incremental,gubela2020response} to create a proxy variable, $Z$, whose expected value corresponds to the causal quantity of interest. In our case, we can define a transformed outcome variable $Z_{ij}$ for each unit $(i,j)$ as:

\begin{equation}
    Z_{ij} =  \begin{cases} +\frac{\overline{Y}_i}{e_{j}(1|X_{i})} & \text{if } A_{ij}=1 \\ -\frac{\overline{Y}_i}{e_{j}(0|X_{i})} & \text{if } A_{ij}=0 \end{cases}
    \label{eq:z_transform}
\end{equation}

Here, the outcome is reweighted by the inverse propensity score. Training a model $f_\theta(X_{ij})$ to predict this transformed variable $Z_{ij}$ by maximizing $\sum_{i,j} Z_{ij} \cdot f_\theta(X_{ij})$ is equivalent to maximizing the AddIPW policy-dependent objective from Equation \ref{eq:general_objective}.

The crucial distinction between our approach and standard response transformations is the use of the cluster-level outcome ($\overline{Y}_i$) instead of the individual outcome ($Y_{ij}$). This modification makes the transformation interference-aware. By using the average outcome of the entire cluster, we evaluate each item's treatment based on its total impact, thus accounting for effects like cannibalization.

This flexible framework allows us to adapt proven economic uplift methods to the clustered network interference setting by simply substituting the cluster-level target value used in Equation \ref{eq:z_transform}. We propose the following adaptations, where $w_{ij} = \left(\frac{A_{ij}}{e_{j}(1|X_{i})} - \frac{1-A_{ij}}{e_{j}(0|X_{i})}\right)$:

\begin{itemize}
    \item \textbf{AddIPW Naive Profit}: A direct approach where the outcome is the average cluster profit, $\overline{Y}_{p,i}$. The model learns to maximize:
    \begin{equation}
        \frac{1}{n}\sum_{i=1}^{n}\overline{Y}_{p,i}\sum_{j=1}^{M_{i}}w_{ij}f_\theta(X_{ij})
    \end{equation}

    \item \textbf{AddIPW CRVTW Adaptation}: We adapt the Continuous Response Variable Transformation with Weightings method~\cite{gubela2020response}. In our adaptation, we simply substitute the conversion outcome in the objective with the average cluster-level revenue $\overline{Y}_{r,i}$. The model maximizes:
    \begin{equation}
        \frac{1}{n}\sum_{i=1}^{n}\overline{Y}_{r,i}\sum_{j=1}^{M_{i}}w_{ij}f_\theta(X_{ij})
    \end{equation}
    
    \item \textbf{AddIPW IPC Adaptation}: The Incremental Profit per Conversion method~\cite{proencca2023incremental} is adapted by using only clusters with at least one conversion ($\sum_j Y_{ij} > 0$). The outcome is the average cluster-level profit, assuming the treatment cost for all converted individuals $\overline{Y}_{c^{T=1},i})$ was applied. The model maximizes:
    \begin{equation}
        \frac{1}{n}\sum_{i=1}^{n} \mathbb{I}\{\sum_{j=1}^{M_i} Y_{ij} > 0\} \cdot (\overline{Y}_{r,i} - \overline{Y}_{c^{T=1},i})) \sum_{j=1}^{M_{i}} w_{ij} f_\theta(X_{ij})
    \end{equation}
\end{itemize}

\section{Experimental results}
For our experiments, we use the simulation framework introduced by \cite{karlsson2025qini}\footnote{https://github.com/bookingcom/uplift-interference-simulator}. We refer to their work for a more detailed explanation of the simulator mechanics. For our experiments, we use the softmax-exponential-decay interference model and set the softmax temperature to $5.0$. Our experiments use $100,000$ samples with clusters of varying sizes. In our experiments, a treatment is equivalent to applying a discount of 8\% to an item.  

We compare each policy on its effectiveness in driving incremental conversion and incremental profit/loss per incremental conversion. 

In addition to the proposed AddIPW-CRVTW and AddIPW-IPC adaptations, we also show the performance of their vanilla (inference-unaware) variants, a random policy, AddIPW (optimized for conversion), and AddIPW directly optimised for $\overline{Y}_{p,i}$ (AddIPW naive profit).

\begin{figure}
    \centering
    \includegraphics[width=1.0\linewidth]{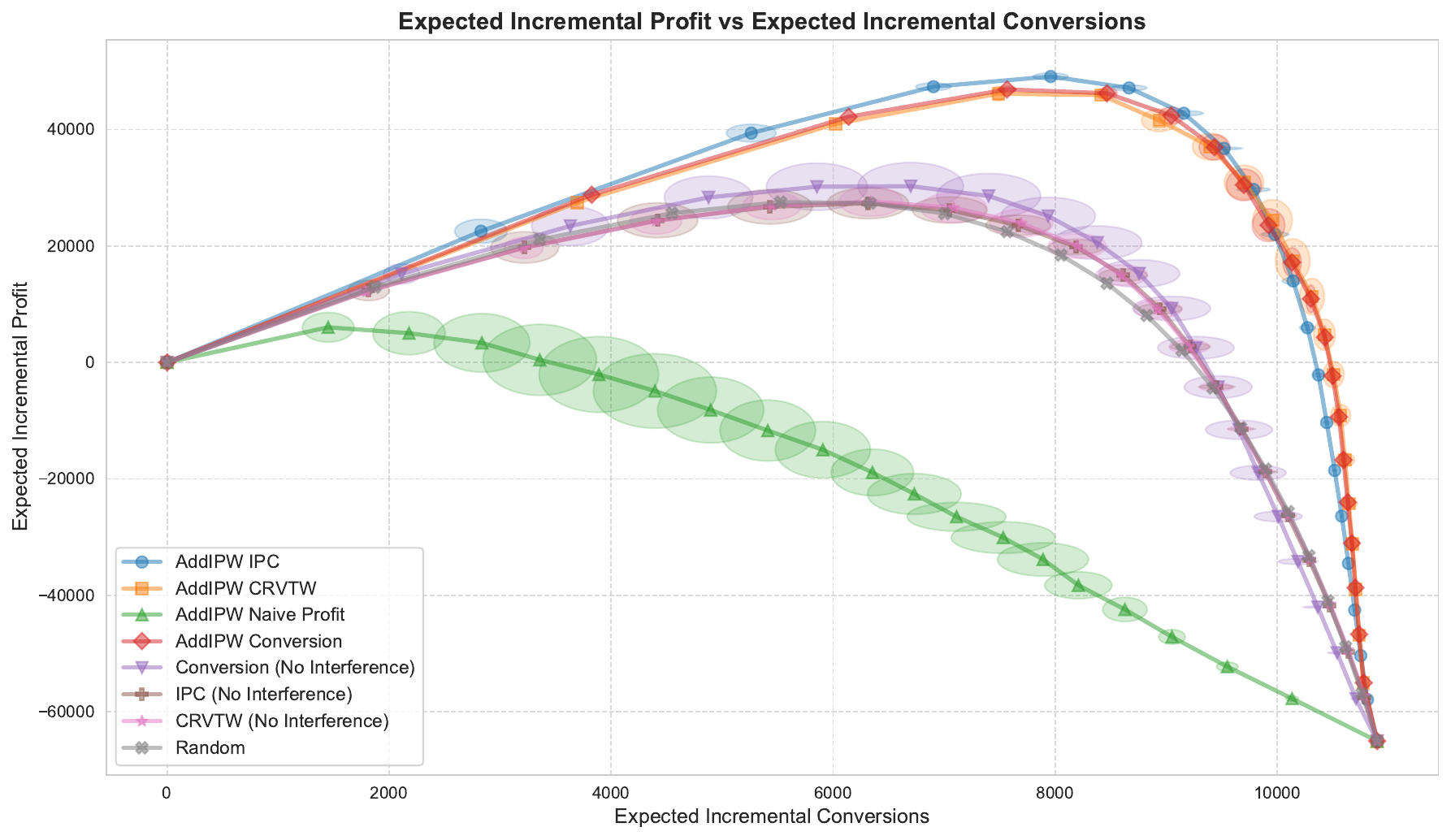}
    \caption{Profit vs. Conversion Qini Curve for models trained on a cluster size of 40. The AddIPW-IPC adaptation shows superior profit generation at lower treatment fractions (i.e., when targeting the highest-uplift users). Error bars represent one standard deviation.}
    \label{fig:full-qini-curve}
\end{figure}

The performance of the various models is evaluated using Qini curves~\cite{radcliffe2007using}, presented in Figures \ref{fig:aucvsclusters} and \ref{fig:full-qini-curve}. Each point on a curve corresponds to a policy of targeting a specific fraction of the population, $\phi$, sorted in descending order by their predicted uplift scores. The points are plotted for cumulative population fractions in 5\% increments (i.e., for $\phi=0.0, 0.05,0.10,\ldots,1.0$).

In Figure~\ref{fig:aucvsclusters}, we show that as the size of clusters grows, the performance gap between the interference-naive and AddIPW methods grows as well. In contrast, the difference between directly optimizing conversion, IPC, and CRVTW is minimal, with AddIPW-IPC gaining a small yet significant advantage over the other two methods. However, in Figure~\ref{fig:full-qini-curve} we see AddIPW-IPC has a larger advantage in the lower treatment percentages. In Figure~\ref{fig:auc70vscluster}, we show that by focusing on the top 70\% of prioritized items, the AddIPW-IPC method demonstrates a more pronounced advantage. This is especially interesting, as AddIPW-IPC uses far less data compared to the other AddIPW methods as it drops non-converted clusters.    

It is worth noting that in our preliminary experiments, the AddIPW approach did not universally outperform naive methods at lower softmax temperatures. We assume this phenomenon is attributable to the diminished strength of interference under these conditions, which may provide an advantage to the lower-variance~\cite{karlsson2025qini} naive methods.

\section{Conclusion}
This paper addresses the challenge of learning uplift models in the presence of clustered network interference, a common scenario that violates the SUTVA assumptions made by standard uplift modeling methods. We proposed a practical methodology that leverages AddIPW, an existing interference-aware policy estimator, as a differentiable objective function.

Our experiments demonstrate two key findings. First, interference-aware policies learned using our AddIPW-based objective substantially outperform traditional, interference-naive models, with the performance advantage growing as the degree of interference increases. Second, by integrating established economic uplift strategies like IPC into the AddIPW framework, we can create policies that optimize for profit directly. This is a crucial advantage for applications with limited budgets where prioritizing the highest-impact treatments is paramount.
This work provides a foundational framework for developing and optimizing a new class of interference-aware uplift models. It opens a clear path for future research to adapt more sophisticated uplift techniques and to validate these methods in complex, large-scale production environments.

\begin{acks}
We thank Rickard Karlsson, Felipe Moraes, Itsik Adiv, and Ulf Schnabel for their insightful discussions and feedback.
\end{acks}

\bibliographystyle{ACM-Reference-Format}
\bibliography{main}


\begin{thebibliography}{10}


\ifx \showCODEN    \undefined \def \showCODEN     #1{\unskip}     \fi
\ifx \showISBNx    \undefined \def \showISBNx     #1{\unskip}     \fi
\ifx \showISBNxiii \undefined \def \showISBNxiii  #1{\unskip}     \fi
\ifx \showISSN     \undefined \def \showISSN      #1{\unskip}     \fi
\ifx \showLCCN     \undefined \def \showLCCN      #1{\unskip}     \fi
\ifx \shownote     \undefined \def \shownote      #1{#1}          \fi
\ifx \showarticletitle \undefined \def \showarticletitle #1{#1}   \fi
\ifx \showURL      \undefined \def \showURL       {\relax}        \fi
\providecommand\bibfield[2]{#2}
\providecommand\bibinfo[2]{#2}
\providecommand\natexlab[1]{#1}
\providecommand\showeprint[2][]{arXiv:#2}

\bibitem[Agarwal et~al\mbox{.}(2019)]%
        {agarwal2019general}
\bibfield{author}{\bibinfo{person}{Aman Agarwal}, \bibinfo{person}{Kenta Takatsu}, \bibinfo{person}{Ivan Zaitsev}, {and} \bibinfo{person}{Thorsten Joachims}.} \bibinfo{year}{2019}\natexlab{}.
\newblock \showarticletitle{A general framework for counterfactual learning-to-rank}. In \bibinfo{booktitle}{\emph{Proceedings of the 42nd International ACM SIGIR Conference on Research and Development in Information Retrieval}}. \bibinfo{pages}{5--14}.
\newblock


\bibitem[Gubela et~al\mbox{.}(2020)]%
        {gubela2020response}
\bibfield{author}{\bibinfo{person}{Robin~M Gubela}, \bibinfo{person}{Stefan Lessmann}, {and} \bibinfo{person}{Szymon Jaroszewicz}.} \bibinfo{year}{2020}\natexlab{}.
\newblock \showarticletitle{Response transformation and profit decomposition for revenue uplift modeling}.
\newblock \bibinfo{journal}{\emph{European Journal of Operational Research}} \bibinfo{volume}{283}, \bibinfo{number}{2} (\bibinfo{year}{2020}), \bibinfo{pages}{647--661}.
\newblock


\bibitem[Hudgens and Halloran(2008)]%
        {hudgens2008toward}
\bibfield{author}{\bibinfo{person}{Michael~G Hudgens} {and} \bibinfo{person}{M~Elizabeth Halloran}.} \bibinfo{year}{2008}\natexlab{}.
\newblock \showarticletitle{Toward causal inference with interference}.
\newblock \bibinfo{journal}{\emph{Journal of the american statistical association}} \bibinfo{volume}{103}, \bibinfo{number}{482} (\bibinfo{year}{2008}), \bibinfo{pages}{832--842}.
\newblock


\bibitem[Karlsson et~al\mbox{.}(2025)]%
        {karlsson2025qini}
\bibfield{author}{\bibinfo{person}{Rickard~KA Karlsson}, \bibinfo{person}{Bram van~den Akker}, \bibinfo{person}{Felipe Moraes}, \bibinfo{person}{Hugo~M Proen{\c{c}}a}, {and} \bibinfo{person}{Jesse~H Krijthe}.} \bibinfo{year}{2025}\natexlab{}.
\newblock \showarticletitle{Qini curve estimation under clustered network interference}.
\newblock \bibinfo{journal}{\emph{arXiv preprint arXiv:2502.20097}} (\bibinfo{year}{2025}).
\newblock


\bibitem[Oosterhuis and de~Rijke(2021)]%
        {oosterhuis2021unifying}
\bibfield{author}{\bibinfo{person}{Harrie Oosterhuis} {and} \bibinfo{person}{Maarten de Rijke}.} \bibinfo{year}{2021}\natexlab{}.
\newblock \showarticletitle{Unifying online and counterfactual learning to rank: A novel counterfactual estimator that effectively utilizes online interventions}. In \bibinfo{booktitle}{\emph{Proceedings of the 14th ACM international conference on web search and data mining}}. \bibinfo{pages}{463--471}.
\newblock


\bibitem[Proen{\c{c}}a and Moraes(2023)]%
        {proencca2023incremental}
\bibfield{author}{\bibinfo{person}{Hugo~Manuel Proen{\c{c}}a} {and} \bibinfo{person}{Felipe Moraes}.} \bibinfo{year}{2023}\natexlab{}.
\newblock \showarticletitle{Incremental profit per conversion: a response transformation for uplift modeling in e-commerce promotions}.
\newblock \bibinfo{journal}{\emph{arXiv preprint arXiv:2306.13759}} (\bibinfo{year}{2023}).
\newblock


\bibitem[Radcliffe(2007)]%
        {radcliffe2007using}
\bibfield{author}{\bibinfo{person}{Nicholas Radcliffe}.} \bibinfo{year}{2007}\natexlab{}.
\newblock \showarticletitle{Using control groups to target on predicted lift: Building and assessing uplift model}.
\newblock \bibinfo{journal}{\emph{Direct Marketing Analytics Journal}} (\bibinfo{year}{2007}), \bibinfo{pages}{14--21}.
\newblock


\bibitem[Rubin(1980)]%
        {rubin1980randomization}
\bibfield{author}{\bibinfo{person}{Donald~B Rubin}.} \bibinfo{year}{1980}\natexlab{}.
\newblock \showarticletitle{Randomization analysis of experimental data: The Fisher randomization test comment}.
\newblock \bibinfo{journal}{\emph{Journal of the American statistical association}} \bibinfo{volume}{75}, \bibinfo{number}{371} (\bibinfo{year}{1980}), \bibinfo{pages}{591--593}.
\newblock


\bibitem[Zhang and Imai(2023)]%
        {zhang2023individualized}
\bibfield{author}{\bibinfo{person}{Yi Zhang} {and} \bibinfo{person}{Kosuke Imai}.} \bibinfo{year}{2023}\natexlab{}.
\newblock \showarticletitle{Individualized policy evaluation and learning under clustered network interference}.
\newblock \bibinfo{journal}{\emph{arXiv preprint arXiv:2311.02467}} (\bibinfo{year}{2023}).
\newblock


\bibitem[Zhou et~al\mbox{.}(2023)]%
        {zhou2023direct}
\bibfield{author}{\bibinfo{person}{Hao Zhou}, \bibinfo{person}{Shaoming Li}, \bibinfo{person}{Guibin Jiang}, \bibinfo{person}{Jiaqi Zheng}, {and} \bibinfo{person}{Dong Wang}.} \bibinfo{year}{2023}\natexlab{}.
\newblock \showarticletitle{Direct heterogeneous causal learning for resource allocation problems in marketing}. In \bibinfo{booktitle}{\emph{Proceedings of the AAAI Conference on Artificial Intelligence}}, Vol.~\bibinfo{volume}{37}. \bibinfo{pages}{5446--5454}.
\newblock


\end{thebibliography}



\end{document}